\documentclass[10pt,twocolumn,letterpaper]{article}

\usepackage{cvpr}
\usepackage{times}
\usepackage{epsfig}
\usepackage{graphicx}
\usepackage{amsmath}
\usepackage{amssymb}
\usepackage{algorithm}
\usepackage{algorithmic}
\usepackage{subfigure}
\usepackage{marvosym}
\newcommand{\D}{\mathcal{D}}

\newcommand{\tabincell}[2]{\begin{tabular}{@{}#1@{}}#2\end{tabular}}

\usepackage[breaklinks=true,bookmarks=false]{hyperref}

\cvprfinalcopy 

\def\cvprPaperID{****} 

\ifcvprfinal\fi
\begin{document}

\title{Model Adaptation: Unsupervised Domain Adaptation without Source Data}

\author{Rui Li\textsuperscript{1}, Qianfen Jiao\textsuperscript{1}, Wenming Cao\textsuperscript{3}, Hau-San Wong\textsuperscript{1}, Si Wu\textsuperscript{2}\\
\textsuperscript{1}Department of Computer Science, City University of Hong Kong\\
\textsuperscript{2}School of Computer Science and Engineering, South China University of Technology\\
\textsuperscript{3}Department of Statistics and Actuarial Science, The University of Hong Kong\\
{\tt\small ruili52-c@my.cityu.edu.hk, qjiao4-c@my.cityu.edu.hk, wmingcao@hku.hk} \\
{\tt\small \textsuperscript{\Letter}cshswong@cityu.edu.hk, cswusi@scut.edu.cn}}


\maketitle
\thispagestyle{empty}

\begin{abstract}
In this paper, we investigate a challenging unsupervised domain adaptation setting --- unsupervised model adaptation. We aim to explore how to rely only on unlabeled target data to improve performance of an existing source prediction model on the target domain, since labeled source data may not be available in some real-world scenarios due to data privacy issues. For this purpose, we propose a new framework, which is referred to as collaborative class conditional generative adversarial net to bypass the dependence on the source data. Specifically, the prediction model is to be improved through generated target-style data, which provides more accurate guidance for the generator. As a result, the generator and the prediction model can collaborate with each other without source data. Furthermore, due to the lack of supervision from source data, we propose a weight constraint that encourages similarity to the source model. A clustering-based regularization is also introduced to produce more discriminative features in the target domain. Compared to conventional domain adaptation methods, our model achieves superior performance on multiple adaptation tasks with only unlabeled target data, which verifies its effectiveness in this challenging setting.
\end{abstract}

\section{Introduction}
\begin{figure}[t]
\begin{center}
\includegraphics[width=0.96\linewidth]{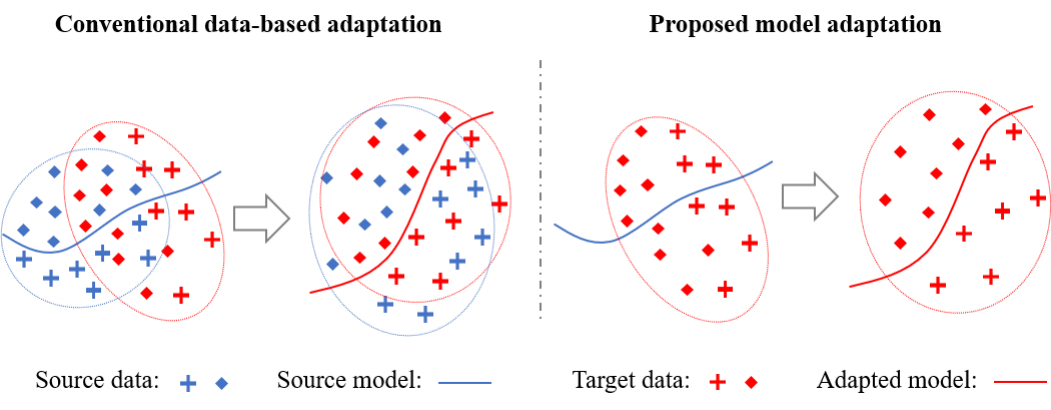}
\end{center}
   \caption{Comparison between conventional data-based adaptation (left) and our model adaptation (right). Conventional unsupervised domain adaptation methods require labeled source data during adaptation, while our proposed model adaptation method only relies on unlabeled target data.}
\label{fig:domain_adaptation_problem}
\end{figure}

Although deep neural networks have achieved state-of-the-art performance on various visual recognition tasks~\cite{AlexNet,Resnet}, the promising performance heavily relies on the availability of sufficient labeled dataset with diverse visual variations~\cite{ImageNet}, and the training and test data should be independent and identically distributed. When the test environment is different from the source domain, the performance of most visual systems will be seriously degraded. This is known as the domain shift~\cite{SurveyTransferLearning}, as illustrated in Fig.~\ref{fig:domain_adaptation_problem}. Domain shift is one of the key factors that prevent the transfer of research results into real-world applications. An intuitive strategy is to re-collect and annotate sufficient target dataset to re-train or fine-tune the model~\cite{TransferableFeatures,LabelEfficientLearning}. However, this solution is not only expensive but also not practical for manual annotation in various environments.

There are great interests in developing a visual recognition model that generalize well to different domains with few or no human annotations. Recently, unsupervised domain adaptation has received a lot of attention, since it makes large progresses in generalizing the pre-trained prediction model to the target domain where labels are not available. This is achieved by exploiting knowledge from sufficiently labeled source data~\cite{SimultaneousAcrossDomainsTasks,DeepRC,JAN}. Existing unsupervised domain adaptation methods normally assume that the source dataset is available during training. However, this assumption is not always practical in the following cases:
\vspace{-0.6em}
\begin{enumerate}
\item For many companies, they will only provide the learned models instead of their customer data due to data privacy and security issues.
\vspace{-0.6em}
\item The source datasets like videos or high-resolution images may be so large that it is not practical or convenient to transfer or retain them to different platforms.
\end{enumerate}
\vspace{-0.6em}

Therefore, developing an unsupervised domain adaptation method without source dataset has a high practical value~\cite{DAnosource}. Recent domain adaptation methods are categorized into two groups: 1) learning domain-invariant features by minimizing a specific distribution distance between the source and target domains~\cite{DAN}; and 2) translating the source data to the target data directly based on generative adversarial networks (GAN)~\cite{GAN}. Despite their great progresses, these methods cannot handle the setting where the source dataset is not available since estimating the source distribution or transformation between two domains is impossible.

In this paper, we focus on unsupervised domain adaptation without source data, referred to as unsupervised model adaptation, and follow the standard assumption where both domains share the same label space. The idea of model-based unsupervised domain adaptation is shown in the right of Fig.~\ref{fig:domain_adaptation_problem}. Specifically, conventional data-based unsupervised domain adaptation aims to learn a prediction model $C$ to generalize to the target domain based on labeled source data $\D_s = \{X_s, Y_s\}$ and unlabeled target data $\D_t = \{X_t\}$, while our model-based adaptation is to adapt the pre-trained source model $C$ to the target domain only with $\D_t$. In other words, $\D_s$ is not accessible during the model adaptation. It is noteworthy that we can easily obtain the pre-trained $C$ with $\D_s$. However, this process cannot be reversed. Therefore, our new model-based adaptation is designed for above scenarios, which adapts an existing model to new domains.

First, to be independent of source data, we develop a Collaborative Class Conditional Generative Adversarial Networks (3C-GAN) for producing target-style training samples. To this end, a discriminator is introduced to match the target distribution by adversarial training. A class conditional generator is imposed with a semantic similarity constraint, which collaborates with the prediction model during the adaptation. Second, we introduce a weight regularization that encourages the prediction model to be close to the original source model, which can stabilize training and improve the performance. Moreover, a clustering-based regularization is incorporated into the overall objective to force the decision boundaries to be located in the low-density regions, thereby improving the final adaptation performance.

We conduct extensive experiments on multiple unsupervised domain adaptation benchmarks. In addition, ablation studies are performed to analyze contributions of each component in our model. The experimental results have verified the superiority of our method. We summarize contributions of this work as follows:
\vspace{-0.6em}
\begin{itemize}
\item We consider a novel and challenging adaptation setting which aims to transfer a prediction model across different domains with only unlabeled data. This is not feasible for the existing adaptation approaches.
\vspace{-0.6em}
\item To avoid relying on source data, we propose 3C-GAN where the generator and the prediction model can be collaboratively enhanced during adaptation.
\vspace{-0.6em}
\item We demonstrate that the proposed model is sufficiently effective on multiple domain adaptation benchmarks, and outperforms recent state-of-the-art results in the absence of source data.
\end{itemize}
\vspace{-0.6em}

\section{Related Work}\label{section:related_work}
In this section, we focus on recent unsupervised domain adaptation methods based on Convolutional Neural Networks (CNNs) due to its superior performance.

Most domain adaptation methods mitigate the distribution discrepancy between domains according to~\cite{TheoryDA}. The expected error on the target domain is bounded by: 1) the expected error on the source domain; 2) the domain discrepancy between the source and target domains; and 3) a shared expected loss which is expected to be small~\cite{LearnSemanticRepresentation}. The expected error on the source domain can be minimized by using labeled data in the source domain. Thus the core task becomes to minimize the discrepancy between domains. 
Deep Domain Confusion (DDC)~\cite{DDC} and Deep Adaptation Networks (DAN)~\cite{DAN} adopt maximum mean discrepancy~\cite{KernelMMD} on the final multiple layers to enforce the distribution similarity between source and target features. Joint Adaptation Networks (JAN)~\cite{JAN} uses the joint maximum mean discrepancy to align the joint distributions among multiple layers. Deep CORAL~\cite{DeepCoral} use feature covariance to measure the domain discrepancy. Philip \textit{et al.}~\cite{AssocDA} enforce the associations of similar features within two domains. In addition to these methods of measuring distribution discrepancy, maximizing the domain confusion via adversarial training can be used to align distributions. Domain Adversarial Neural Network (DANN)~\cite{DANN} introduces a domain classifier and renders the extracted features from two domains indistinguishable by a gradient reversal layer~\cite{RevGral}. These adversarial training based methods show effective adaptation performances~\cite{DSN,CondAdvDA}. Pinheiro \textit{et al.} include the adversarial loss and a similarity-based classifier~\cite{SimilarityUDA} to improve the model generalization. To integrate category information into the learning of domain-invariant features, Multi-Adversarial Domain Adaptation (MADA)~\cite{MADA} adopts multiple domain discriminators which correspond to each category. In~\cite{MCD}, instead of relying on a domain discriminator, Saito \textit{et al.} propose two task classifiers to align distributions by minimizing their discrepancy.~\cite{SWD} adopts sliced Wasserstein metric to measure the dissimilarity of classifiers.

Inspired by GAN~\cite{GAN}, recent works achieve feature distribution alignment based on a generative model. Sankaranarayanan \textit{et al.} propose a GenerateToAdapt model~\cite{GenToAdapt} which induces the extracted source or target embeddings to produce source-like images, such that the extracted features are expected to be domain-invariant. DuplexGAN~\cite{DuplexGAN} uses two discriminators for two domains to ensure that the extracted features can generate images on both domains based on a domain code. Image-to-image translation~\cite{Pix2Pix} provides a new direction for domain adaptation, which achieves the distribution alignment in the data space. In the absence of paired domain data, preserving the content will be non-trivial, and several recent works perform unsupervised image-to-image translation by including an extra constraint between input and the transformed output. SimGAN~\cite{SimGAN} employs a reconstruction loss between them, while PixelDA~\cite{PixelDA} and DTN~\cite{DTN} encourage the output to have the same class label and the semantic features as input, respectively. CoGAN~\cite{CoGAN} and UNIT~\cite{UNIT} learn a feature space based on shared or non-shared strategies to perform cross-domain generation. Zhu \textit{et al.} propose CycleGAN~\cite{CycleGAN} which involves bi-directional translations with a cycle-consistency loss, which enforces the condition that the translated image can be mapped back to input. DiscoGAN~\cite{DiscoGAN} and DualGAN~\cite{DualGAN} share the same idea and achieve promising unsupervised image translation performance. CyCADA~\cite{CYCADA} is based on CycleGAN and delivers good performance on multiple domain adaptation tasks.

Additionally, some works further explore using unlabeled target data to improve generalization by co-training~\cite{colearn}, pseudo-labeling~\cite{AsymTriTraining,CRST}, and entropy regularization~\cite{DIRT-T}. Some recent works focus on open set adaptation problems~\cite{UniversalDA}. However, these works require source data during adaptation. Thus, most previous works are not applicable to the proposed model adaptation problem. Some incremental learning works~\cite{LearningWoMemorize,LearningWoForget} are relevant to us, but they need labeled target data for new tasks. In this paper, we propose to simply use the unlabeled target dataset to adapt the pre-trained model to the target domain.

\section{Proposed Method}\label{section:proposed_method}

\begin{figure*}[t]
\centering
\includegraphics[width=0.86\textwidth]{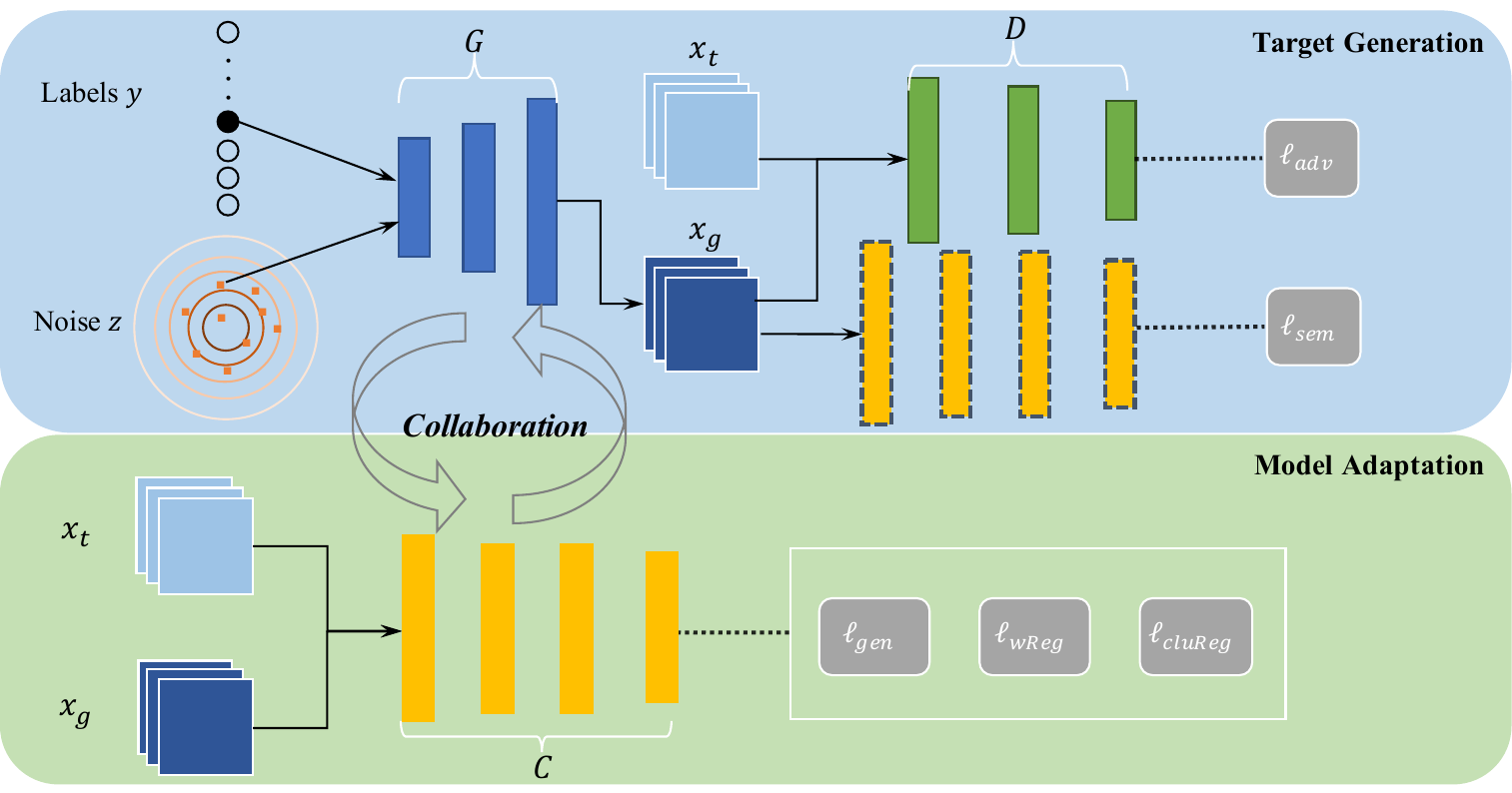}
\caption{An overview of the proposed architecture. During target generation (top), we aim to learn a class conditional generator $G$ for producing target-style training samples $x_g = G(y,z)$ via the discriminator $D$ and the prediction model $C$ (which is fixed as denoted by dashline). The generated images and proposed regularizations are used for model adaptation (bottom). These two procedures are repeated, with $G$ and $C$ collaborating with each other. (See text for details)}
\label{fig:overview}
\end{figure*}

In this section, we elaborate our model for unsupervised model adaptation problem, where we merely have access to the pre-trained prediction model $C$ from the source domain and unlabeled target dataset $X_t$. Our goal is to adapt $C$ to the target domain with $X_t$.

To this end, we propose a Collaborative Class Conditional Generative Adversarial Networks (3C-GAN) for model adaptation in absence of source data. Except for the existing pre-trained $C$, our framework consists of another two components: a discriminator $D$ for matching target distribution and a generator $G$ conditioned on randomly sampled labels for producing valid target-style training samples. By incorporating the generated data during training, the performance of $C$ is improved on the target domain which can in turn promote the generation process of $G$. Besides, we design two regularization terms to prevent the adapted model far away from the pre-trained source model and improve the generalization on the target domain, respectively. The architecture is illustrated in Fig.~\ref{fig:overview}. $D$, $G$ and $C$ are parameterized by $\theta_D$, $\theta_G$ and $\theta_C$, respectively. The details for each proposed component are introduced as follows.

\subsection{Collaborative Class Conditional GAN} \label{subsection:class_conditional_generation}

To avoid using source data for domain adaptation, we propose the Collaborative Class Conditional GAN (3C-GAN) for collaboratively improving the generator $G$ and the prediction model $C$. As shown in Fig.~\ref{fig:overview}, this is achieved by integrating $C$ into the GAN framework. Different from standard GAN model, where $G$ is only conditioned on a noise vector $z$, our $G$ is further conditioned on a pre-defined label $y$, \textit{i.e.}, $x_g = G(y, z)$. Also in contrast to traditional conditional GAN~\cite{ConditionalGAN} where $D$ is trained to distinguish real and fake pairs in a supervised manner, our $D$ is optimized to distinguish $x_t$ from $x_g$. The objective function for $D$ can be expressed as follows:
{\small
\begin{align}
\max_{\theta_D}~\mathbb{E}_{x_t\sim \D_t}[\log D(x_t)] +\mathbb{E}_{y, z}[\log(1 - D(G(y,z)))].
\label{eq:loss_D}
\end{align}
}
Meanwhile, $G$ is updated to fool $D$ by generating $x_g$ with a similar distribution as $x_t$. Thus, the adversarial loss $\ell_{adv}$ of $G$ can be formulated as follows:
\begin{align}
\ell_{adv}(G) = \mathbb{E}_{y, z}[\log D(1 - G(y,z))].
\label{eq:l_adv}
\end{align}
Although $\ell_{adv}$ simulates the target distribution, it cannot guarantee the semantic similarity to the input label $y$.

Inspired by~\cite{InfoGAN}, we propose a semantic similarity loss $\ell_{sem}$ based on the existing prediction model $C$. It enforces the semantic similarity between $x_g$ and the input label $y$ based on the prediction model $C$, as defined below:
\begin{align}
\ell_{sem}(G) = \mathbb{E}_{y,z} [-y\log p_{\theta_C}(G(y,z))],
\label{eq:l_sem}
\end{align}
where $p_{\theta_C}(\cdot)$ indicates the class probability predicted by the prediction model $C$. $\ell_{sem}$ enables the generation semantics. After including $\ell_{adv}$ that matches the target distribution, the optimization objective of generator $G$ is defined as follows:
\begin{align}
\min_{\theta_G} ~ \ell_{adv} + \lambda_{s}\ell_{sem},
\label{eq:loss_G}
\end{align}
where $\lambda_{s}$ balances two losses. We alternately update $D$ and $G$ for optimizing Eq.~(\ref{eq:loss_D}) and Eq.~(\ref{eq:loss_G}), respectively. As a result, $G$ can produce new target-style instances, \textit{i.e.}, $\{x_g, y\}$, which are used to improve performance of $C$ on the target domain. $C$ and $G$ collaborate with each other during training since the enhanced $C$ can provide more accurate guidance for $G$ and a more reliable generation can in turn improve performance of $C$. Therefore, the overall framework refers to collaborative class conditional generative adversarial networks.

In addition to $\ell_{gen} = \mathbb{E}_{y,z} [-y\log p_{\theta_C}(x_g)]$, we further include two regularizations to enhance the performance of $C$. The final optimization objective for the prediction model $C$ can be expressed as below:
\begin{align}
\min_{\theta_C} ~ \lambda_g \ell_{gen} + \lambda_w \ell_{wReg} + \lambda_{clu} \ell_{cluReg},
\label{eq:loss_C}
\end{align}
where $\ell_{wReg}$ and $\ell_{cluReg}$ denote weight regularization and clustering-based regularization. $\lambda_g$, $\lambda_w$ and $\lambda_{clu}$ are used to adjust relative effects of each loss. During the adaptation process, the source dataset is not used, as shown in Fig.~\ref{fig:overview}.

\subsection{Weight Regularization} \label{subsection:w_regularization}

Although only incorporating the above generated target-style samples into training $C$ can improve its performance, the training process is not always stable due to the lack of accurate supervision from the labeled source data. Inspired by~\cite{BeyondSharingWeights,ADDA} which attempt to learn two separate but related prediction models for the source and target domains, we propose a weight regularization term $\ell_{wReg}$ to prevent the parameters of the prediction model $C$ to drift far away from those of the pre-trained model learnt in the source dataset. It can be defined as follows:
\begin{align}
\ell_{wReg} = \left\| \theta_C - \theta_{C_s}   \right\|^2,
\label{eq:l_wReg}
\end{align}
where $\theta_{C_s}$ is the parameters of $C$ pre-trained on the source domain, which is fixed. We can observe that if $\theta_{C_s}$ is set to 0, $\ell_{wReg}$ will be reduced to standard weight decay regularization ($\ell_2$). On one hand, $\ell_{wReg}$ prevents the adapted model from changing too significantly, which is helpful in stabilizing the adaptation. On the other hand, enforcing the adapted model similar to the source model can be regarded as preserving the source knowledge. Experiments have verified that $\ell_{wReg}$ leads to better adaptation in most cases.

\subsection{Clustering-based Regularization} \label{subsection:clu_regularization}

Most domain adaptation methods focus on the adaptation process, where unlabeled real target data are only used to estimate the target distribution, while we consider that unlabeled target data can be used to explore the discriminative information on the target domain. The cluster assumption implies that the decision boundaries of the prediction model should not go through data regions with high density~\cite{SemiEntropyMin}. Therefore, we minimize the conditional entropy of the predicted probability on the target domain, as defined by:
\begin{align}
\mathbb{E}_{x_t\sim \D_t}[-p_{\theta_C}(x_t)\log p_{\theta_C}(x_t)].
\label{eq:l_ent}
\end{align}
However, as pointed out in~\cite{SemiEntropyMin}, the conditional entropy derived in Eq.~(\ref{eq:l_ent}) is not reliable when the prediction model is not locally smooth. To improve the approximation of conditional entropy on unlabeled target data, a local smoothness constraint should be added, which is defined as follows:
\begin{align}
\mathbb{E}_{x_t\sim \D_t}\Big[\max_{\left\| r\right\|\leq\xi} \texttt{KL}( p_{\theta_C}(x_t) || p_{\theta_C}(x_t + r))\Big],
\label{eq:l_smooth}
\end{align}
where $\texttt{KL}(\cdot||\cdot)$ denotes the Kullback-Leibler divergence. Following~\cite{VAT}, we attempt to find a perturbation $r$ that affects the prediction most within an intensity range of $\xi$. This constraint forces the prediction output to be similar between $x_t$ and $x_t+r$. Consequently, the prediction model is locally smooth for each unlabeled target sample.

Therefore, the final clustering-based regularization is formulated as follows:
\begin{equation}
\begin{aligned}
\ell_{cluReg} &=  \mathbb{E}_{x_t\sim \D_t}[-p_{\theta_C}(x_t)\log p_{\theta_C}(x_t)] \\&+ [\texttt{KL}( p_{\theta_C}(x_t) || p_{\theta_C}(x_t + \tilde{r}))],
\end{aligned}
\label{eq:l_cluReg}
\end{equation}
where $\tilde{r}$ is the adversarial perturbation derived from Eq.~(\ref{eq:l_smooth}).

\subsection{Implementation Details}
\begin{algorithm}[h]
\caption{Pseudo-code of our model adaptation process}{\tiny}
\small
\label{alg:TrainingProcedures}
\begin{algorithmic}[1]
 \renewcommand{\algorithmicrequire}{\textbf{Input:}}
 \renewcommand{\algorithmicensure}{\textbf{Output:}}
 \REQUIRE Pre-trained prediction model $C$ on the source domain, unlabeled data $X_t$ in the target domain, $\lambda_{g}$, $\lambda_{clu}$ and $\lambda_{w}$, batch size $B$;
 \ENSURE  $\theta_C$ for the prediction model $C$;
 \\ Initialize learning rates $\zeta_G$, $\zeta_D$ and $\zeta_C$ for $G$, $D$ and $C$;\\
  \FOR {$epoch$ = $1$ to $N$}
  \STATE Randomly sample $x_t$ of size $B$ from $X_t$, and random vectors $\{y, z\}$ from the uniform distribution;
  \FOR {each mini-batch}
    \STATE Generate new samples with $y$ and $z$: $X_g=G(y, z)$
    \STATE Update $D$ via $\theta_D\leftarrow \texttt{Adam}(\nabla_{\theta_D}(\sum\limits_{x_t}\log D(x_t)+\sum\limits_{y,z}\log D(1-G(y,z))),\theta_D,\zeta_D)$.
    \STATE Update $G$ via
     $\theta_G\leftarrow\texttt{Adam}(\nabla_{\theta_G}(\ell_{adv} + \lambda_{s}\ell_{sem}), \theta_G, \zeta_G)$
    \IF{starting adaptation}
    \STATE Update $C$ via
    $\theta_C\leftarrow\texttt{Adam}(\nabla_{\theta_C}(\lambda_g\ell_{gen}+\lambda_w\ell_{wReg}+\lambda_{clu}\ell_{cluReg}),\theta_C,\zeta_C)$
    \ENDIF
  \ENDFOR
  \ENDFOR
 \end{algorithmic}
\end{algorithm}
Learning proceeds by alternately updating $C$, $D$ and $G$ to optimize the corresponding objectives in Eq.~\ref{eq:loss_C}, Eq.~\ref{eq:loss_D} and Eq.~\ref{eq:loss_G}, respectively. In the experiments, we do not apply $\ell_{gen}$ and $\ell_{cluReg}$ for $C$ until the generator can produce meaningful data after several steps. The whole model is trained end-to-end and the implementation is shown in Algorithm~\ref{alg:TrainingProcedures}.

\section{Experiments}\label{section:experiments}

In this section, we conduct extensive experiments on multiple domain adaptation benchmarks to verify the effectiveness of our method. For each task, we only use source data to obtain the pre-trained source model, and it is not used during adaptation. The results of recent state-of-the-art domain adaptation methods are presented for comparisons or as references since most of them are not applicable when source data are not available during adaptation process.

\subsection{Experimental Settings}

\noindent \textbf{Digit and sign datasets:} we evaluate our method among five digit datasets (MNIST~\cite{MNIST}, USPS~\cite{USPS}, MNIST-M~\cite{DANN}, SVHN~\cite{SVHN}, Syn.Digits~\cite{DANN}) and two traffic sign datasets (Syn.Signs~\cite{Synsign} and GTSRB~\cite{GTSRB}). The digit datasets contain 10 shared classes, while the traffic sign datasets contain 43 classes. Besides, Syn.Digits and Syn.Signs are synthetic domains, which is more interesting in real applications.

\noindent \textbf{Office-31~\cite{Office-31}} is a standard domain adaptation benchmark, where images are collected from three distinct domains: Amazon (\textbf{A}), Webcam (\textbf{W}) and DSLR (\textbf{D}). Three domains share 31 classes and contain 2817, 795 and 498 samples, respectively. Following~\cite{MADA,JAN}, we evaluate on all six domain adaptation tasks. These tasks can verify the effectiveness of our method when the number of samples is small.

\noindent \textbf{VisDA17~\cite{VisDA17}} is a challenging dataset for domain adaptation from synthetic domain to real domain with 12 shared classes. The synthetic domain contains around 152k images produced by rendering 3D models under different conditions. We use the validation set as the real domain, which contains around 55k images collected from MSCOCO~\cite{MSCOCO}. Since the number of source data is very large, this task can demonstrate the superiority of our method which can achieve successful adaptation without source data.

For experiments on digit and sign datasets, we resize all images to 32$\times$32$\times$3. The architecture of $C$ is similar to the one in~\cite{DIRT-T} for a fair comparison. An UpResBlock module is adopted in the generator for high-quality image generation. We adopt spectral normalization~\cite{SNGAN} in the discriminator for training stability. For experiments on Office-31 and VisDA17, we choose ResNet50 and ResNet101~\cite{Resnet} pre-trained on ImageNet~\cite{ImageNet} to extract features. Both generator and discriminator consist of two dense layers.

We use Adam~\cite{Adam} to optimize all the networks. The learning rates for $D$ and $G$ are $4\times10^{-4}$ and $10^{-4}$, respectively. As to $C$, the initial learning rates are $10^{-3}$ and $10^{-4}$ for digit/sign datasets and office-31, respectively. We decreased it 10 times during the training. For VisDA17, the learning rate is fixed with $10^{-5}$. The weighting factor $\lambda_{w}$, $\lambda_{g}$ and $\lambda_{clu}$ are set to $10^{-4}$, $10^{-1}$ and $1$, respectively. For digit datasets, $\lambda_{clu}$ is set to $10^{-1}$ instead.

\subsection{Experimental Results}

\begin{table*}[t]
\begin{center}
\resizebox{0.90\textwidth}{!}{
\begin{tabular}[\textwidth]{l c c c c c c}
\hline
Method & SVHN$\rightarrow$MNIST & MNIST$\rightarrow$USPS & USPS$\rightarrow$MNIST & MNIST$\rightarrow$MNIST-M  & Syn.Digits$\rightarrow$SVHN & Syn.Sign$\rightarrow$GTSRB \\
\hline
Source-Only                  & 76.4$\pm1.5$ & 92.4$\pm1.7$ & 86.1$\pm1.3$ & 54.2$\pm0.9$  & 86.2$\pm0.9$ & 78.3$\pm1.6$ \\
DAN~\cite{DAN}               & 71.1         & 81.1         & -            & 76.9  &   88  & 91.1 \\
AssocDA~\cite{AssocDA}       & 97.6         & -            & -            & 89.5  &   91.8    & 97.6 \\
DANN~\cite{DANN}             & 73.8         & 85.1         & 73.0         & 77.4  &   91.1    & 88.7  \\
UNIT~\cite{UNIT}                        & 90.5         & 95.9         & 93.5         & -     & - & -             \\
GenToAdapt~\cite{GenToAdapt}            & 92.4$\pm0.9$ & 95.3$\pm0.7$ & 90.8$\pm1.3$ & -    &-  & -       \\
DSN~\cite{DSN}                          & 82.7         & 91.3         & -           & 83.2  & 91.2 & 93.1 \\
PixelDA~\cite{PixelDA}                  & -            & 95.9         & -       & 98.2  & - & -        \\
CyCADA~\cite{CYCADA}                   & 90.4$\pm0.4$ & 95.6$\pm0.2$ & 96.5$\pm0.1$         & -   & - & -         \\
SimDA~\cite{SimilarityUDA}              & -    &96.4    &95.6       &   90.5  & - & -    \\
MCD~\cite{MCD}                     &  96.2$\pm0.4$    &  94.2$\pm0.7$ &  94.1$\pm0.3$  &-  & - & 94.4$\pm0.3$  \\
VADA~\cite{DIRT-T}                     & 97.9         & -         & -         & 97.7   &   94.8  & 98.8       \\
DIRT-T~\cite{DIRT-T}                    & 99.4        & -            & -         & \textbf{98.9} &  \textbf{96.1} & 99.5 \\
\hline
Our Model                 & \textbf{99.4$\pm0.1$}  &\textbf{97.3$\pm0.2$}  & \textbf{99.3$\pm0.1$}  & 98.5$\pm0.2$  & 95.9$\pm0.2$  & \textbf{99.6$\pm0.1$}  \\
\hline
\end{tabular}}
\caption{Classification accuracy (\%) on digit and sign dataset. `-' denotes that the results are not reported.}
\label{table:digit_sign_Ex}
\end{center}
\end{table*}

\begin{table*}[t]
\small
\begin{center}
\begin{tabular}[\textwidth]{l c c c c c c c}
\hline
Method & A$\rightarrow$W & D$\rightarrow$W & W$\rightarrow$D & A$\rightarrow$D  & D$\rightarrow$A & W$\rightarrow$A & Average \\
\hline
ResNet50~\cite{Resnet}       & 68.4$\pm0.2$ & 96.7$\pm0.1$ & 99.3$\pm0.1$ & 68.9$\pm0.2$  & 65.2$\pm0.3$ & 60.7$\pm0.3$  & 76.1\\
DAN~\cite{DAN}               & 80.5$\pm0.4$ & 97.1$\pm0.2$ & 99.6$\pm0.1$ & 78.6$\pm0.2$ & 63.6$\pm0.3$ & 62.8$\pm0.2$ & 80.4   \\
RTN~\cite{RTN}               & 84.5$\pm0.2$ & 96.8$\pm0.1$ & 99.4$\pm0.1$ & 77.5$\pm0.3$ & 66.2$\pm0.2$ & 64.8$\pm0.3$ & 81.6 \\
DANN~\cite{DANN}             & 82.6$\pm0.4$ & 96.9$\pm0.2$ & 99.3$\pm0.2$ & 81.5$\pm0.4$ & 68.4$\pm0.5$ & 67.5$\pm0.5$ & 82.7  \\
ADDA~\cite{ADDA}             & 86.2$\pm0.5$ & 96.2$\pm0.3$ & 98.4$\pm0.3$ & 77.8$\pm0.3$ & 69.5$\pm0.4$ & 68.9$\pm0.5$ & 82.9 \\
JAN~\cite{JAN}               & 86.0$\pm0.4$ & 96.7$\pm0.3$ & 99.7$\pm0.1$ & 85.1$\pm0.4$ & 69.2$\pm0.4$ & 70.7$\pm0.5$ & 84.6 \\
MADA~\cite{MADA}             & 90.0$\pm0.2$ & 97.4$\pm0.1$ & 99.6$\pm0.1$  &87.8$\pm0.2$  &70.3$\pm0.3$ & 66.4$\pm0.3$ & 85.2 \\
GenToAdapt~\cite{GenToAdapt} & 89.5$\pm0.5$ & 97.9$\pm0.3$ & 99.8$\pm0.2$ & 87.7$\pm0.5$ & 72.8$\pm0.3$ & 71.4$\pm0.4$ & 86.5 \\
\hline
Our Model                 & \textbf{93.7$\pm0.2$}  &\textbf{98.5$\pm0.1$}  & \textbf{99.8$\pm0.2$}  & \textbf{92.7$\pm0.4$}  & \textbf{75.3$\pm0.5$}  & \textbf{77.8$\pm0.1$} & \textbf{89.6} \\
\hline
\end{tabular}
\end{center}
\caption{Classification accuracy (\%) on office-31 based on ResNet50~\cite{Resnet}.}
\label{table:office31_Ex}
\end{table*}

\textbf{Results on digit and sign benchmarks:} Table~\ref{table:digit_sign_Ex} compares the classification accuracy of our model adaptation and recent unsupervised domain adaptation methods. First, compared with the Source-Only model (baseline), the performance of our model on the target domain is significantly increased on all the domain adaptation tasks. In particular, the accuracy rate of our model in MNIST$\rightarrow$MNIST-M can reach 98.5\%, which outperforms the baseline by around 40\%. The significant performance gains suggest that the labeled data on the source domain is not sufficient to achieve good generalization performance on the target domain, while the generated target-style training instances and regularizations in our proposed model facilitate the adaptation and largely improve the performance on the target domain. Second, all the other recent domain adaptation methods require the source data during adaptation process, while our model obtains the best or comparable performance in the absence of source data compared with the other competing methods. Specifically, the test accuracies of our model on the tasks SVHN$\rightarrow$MNIST, USPS$\rightarrow$MNIST and Syn.Sign$\rightarrow$GTSRB are greater than 99\%. On MNIST$\rightarrow$MNIST-M and Syn.Digits$\rightarrow$SVHN, our method obtains 98.5\% and 95.9\%, which are competitive to DIRT-T (98.9\% and 96.1\%). However, DIRT-T is based on VADA which involves source data during the first adaptation stage. Interestingly, we observe that our model can achieve 99.2\% and 96.7\% in terms of accuracy when including source data during training, which outperforms DIRT-T.

\textbf{Results on Office-31:} Table~\ref{table:office31_Ex} shows the performances of our model and the other unsupervised domain adaptation methods. All the results are obtained with ResNet50 as the backbone. The first row shows the performance by fine-tuning on the source domain as the baseline. It is clear that our model outperforms all competing methods by a large margin. Specifically, compared to GenToAdapt~\cite{GenToAdapt} and MADA~\cite{MADA} which involve complex architectures and objective functions, our model boosts performance by around 3\% and 4\% on average across six adaptation tasks. In addition, our model shows superior performance on difficult adaptation tasks, \textit{i.e.}, A$\leftrightarrows$D, A$\leftrightarrows$W. It exceeds the performance of the second best method by 4.5\% on average among these four tasks.
\begin{table*}[t]
\begin{center}
\resizebox{0.96\textwidth}{!}{
\begin{tabular}[\textwidth]{l c c c c c c c c c c c c c}
\hline
Method & plane & bcycl & bus & car  & horse & knife & mcycl & person & plant & sktbrd & train & truck & Average \\
\hline
Source-Only                 & 55.1 & 53.3 & 61.9 & 59.1 & 80.6 & 17.9 & 79.7 & 31.2 & 81.0 & 26.5 & 73.5 & 8.5 & 52.4\\
DAN~\cite{DAN}               & 87.1 & 63.0 & 76.5 & 42.0 & 90.3 & 42.9 & 85.9 & 53.1 & 49.7 & 36.3 & 85.8 & 20.7 & 61.1 \\
MCD~\cite{MCD}               & 87.0 & 60.9 & 83.7 & 64.0 & 88.9 & 79.6 & 84.7 & 76.9 & 88.6 & 40.3 & 83.0 & 25.8 & 71.9 \\
SWD~\cite{SWD}               & 90.8 & 82.5 & 81.7 & 70.5 & 91.7 & 69.5 & 86.3 & 77.5 & 87.4 & 63.6 & 85.6 & 29.2 & 76.4 \\
SimDA~\cite{SimilarityUDA}(ResNet152) & 94.3 & 82.3 & 73.5 & 47.2 & 87.9 & 49.2 & 75.1 & 79.7 & 85.3 & 68.5 & 81.1 & 50.3 & 72.9 \\
Self-Ensembling~\cite{SelfEnsembling} (min aug) & 92.9 & 84.9 & 71.5 & 41.2 & 88.8 & 92.4 & 67.5 & 63.5 & 84.5 & 71.8 & 83.2 & 48.1 & 74.2 \\
\hline
Our Model                    & 94.8 & 73.4 & 68.8 & 74.8 & 93.1 & 95.4 & 88.6 & 84.7 & 89.1 & 84.7 & 83.5 & 48.1 & 81.6 \\
Our Model $\dagger$          & 95.7 & 78.0 & 69.0 & 74.2 & 94.6 & 93.0 & 88.0 & 87.2 & 92.2 & 88.8 & 85.1 & 54.3 & \textbf{83.3} \\
\hline
\end{tabular}}
\end{center}
\caption{Class-wise accuracy (\%) on VisDA17 based on ResNet101~\cite{Resnet}. $\dagger$ denotes that we use an enhanced version of ResNet101 which replaces the first 7$\times$7 convolution with three 3$\times$3 convolutions.}
\label{table:visda17_Ex}
\end{table*}

\textbf{Results on VisDA17:} Table~\ref{table:visda17_Ex} shows the class-level accuracy on VisDA17 based on ResNet101. Our model significantly outperforms other unsupervised domain adaptation methods. Specifically, our model achieves 81.6\% class mean accuracy with the vanilla ResNet101, and this result can be further improved with a more powerful backbone. For example, we use an enhanced ResNet101 shown in the last row of Table~\ref{table:visda17_Ex}. The accuracy is increased to 83.3\%, which surpasses SimDA~\cite{SimilarityUDA} with ResNet152 by 10.4\%. Besides, self-ensembling (SE)~\cite{SelfEnsembling} relies on data augmentation and ensemble techniques, while our model outperforms SE (with minimal augmentation) by 9.1\% without data augmentation. In addition, our model does not use source data during adaptation, which is more preferable in this task when the source dataset is rather large.

\subsection{Visualization Analysis}

\begin{figure}[t]
\centering
    \subfigure[MNIST$\rightarrow$MNIST-M]{
        \includegraphics[width=0.38\linewidth]{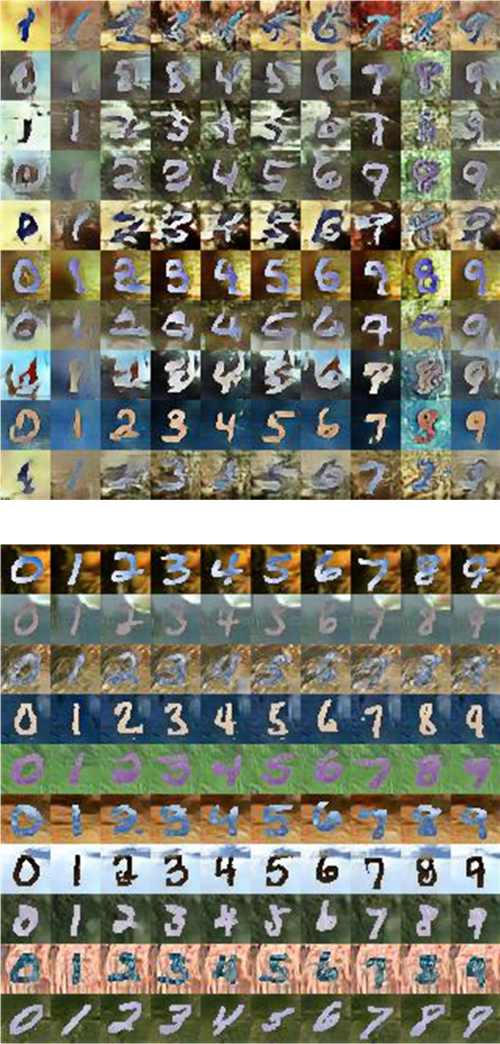}}
    ~~~~~~
    \subfigure[SVHN$\rightarrow$MNIST]{
        \includegraphics[width=0.38\linewidth]{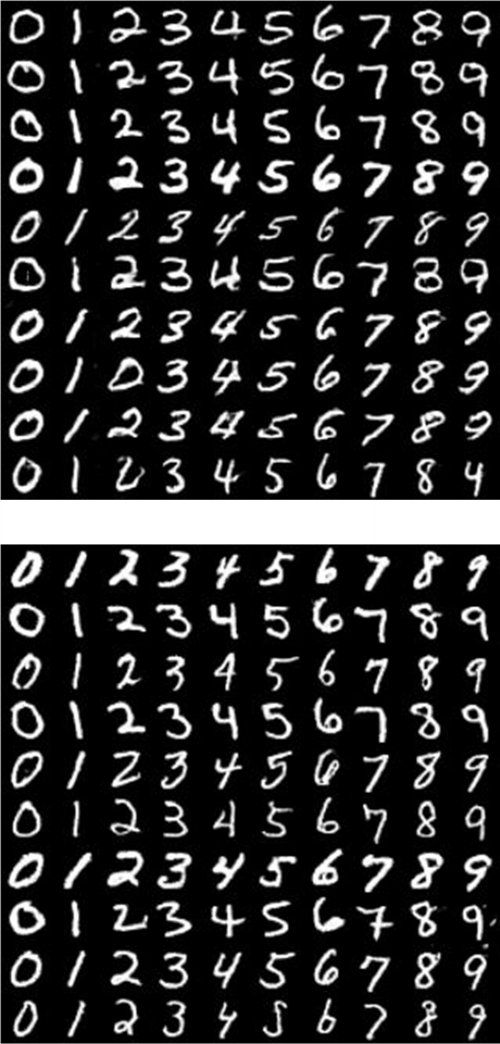}}
\caption{Class conditional generation in (a) MNIST$\rightarrow$MNIST-M and (b) SVHN$\rightarrow$MNIST. The top row indicates the samples generated with pre-trained source model, and the bottom row refers to the samples generated during the last adaptation stage.}
\label{fig:class_conditional_generation_comparisons}
\end{figure}

To provide insights into the collaborative mechanism in our 3C-GAN, we present the generated samples conditioned on the labels from 0 to 9. As shown in Fig.~\ref{fig:class_conditional_generation_comparisons}, each column shares the same class label, and each row shares the same noise vector. Fig.~\ref{fig:class_conditional_generation_comparisons} (top) represents the samples produced in the early stage when $C$ is weak on the target domain, and Fig.~\ref{fig:class_conditional_generation_comparisons} (bottom) represents the samples produced in the late stage of adaptation. We observe that our generator can learn the class-conditional data distribution in these tasks. Besides, after incorporating generated instances into training the prediction model, the performance of the prediction model is increased (see Table~\ref{table:digit_sign_Ex}). The enhanced prediction model can also improve the target class distribution learning within the generator. An obvious illustration is shown in Fig.~\ref{fig:class_conditional_generation_comparisons}(a). The generation quality becomes much better during the late stage when the adapted prediction model is improved on the target domain. It suggests that $C$ and $G$ can collaborate with each other during adaptation process.

\begin{figure}[t]
\centering
    \subfigure[Syn.Digits$\rightarrow$SVHN]{
        \includegraphics[width=0.15\textwidth]{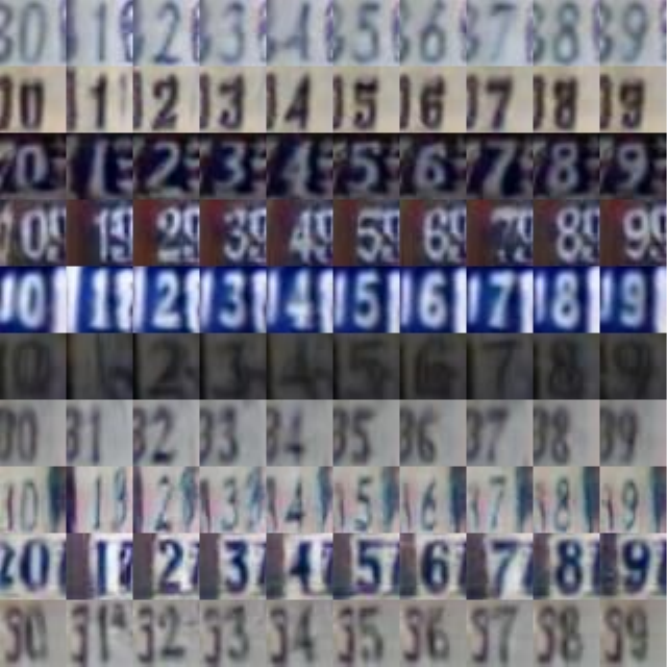}}
    ~~
    \subfigure[Syn.Sign$\rightarrow$GTSRB]{
        \includegraphics[width=0.285\textwidth]{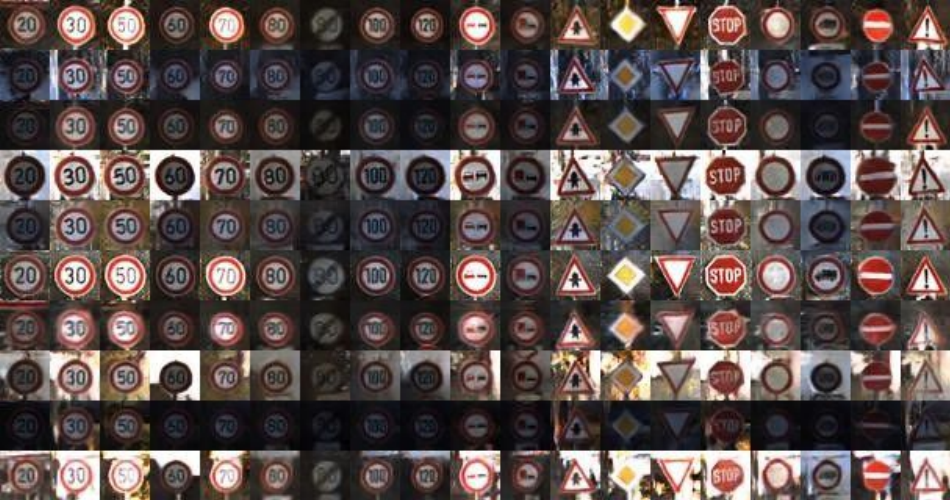}}
\caption{Class-conditional generation in (a) Syn.Digits$\rightarrow$SVHN and (b) Syn.Sign$\rightarrow$GTSRB (shows the first 19 out of 43 classes). Each column has the same class $y$ and the rows share the same noise vector $z$.}
\label{fig:class_conditional_generations_synthetic2real}
\end{figure}

\begin{figure}[t]
\centering
    \subfigure[Before Adaptation]{
        \includegraphics[width=0.20\textwidth]{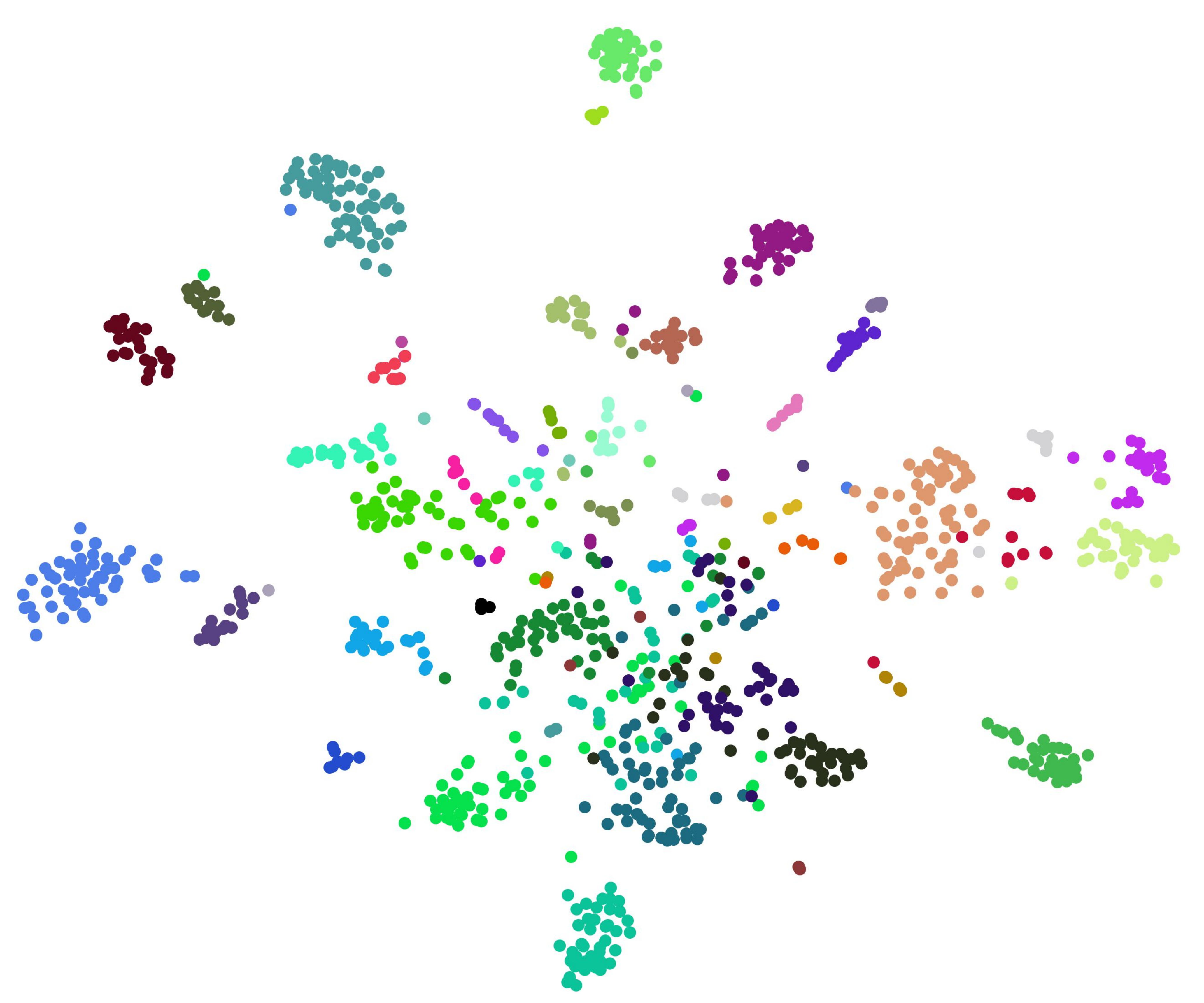}}
    ~~
    \subfigure[After Adaptation]{
        \includegraphics[width=0.20\textwidth]{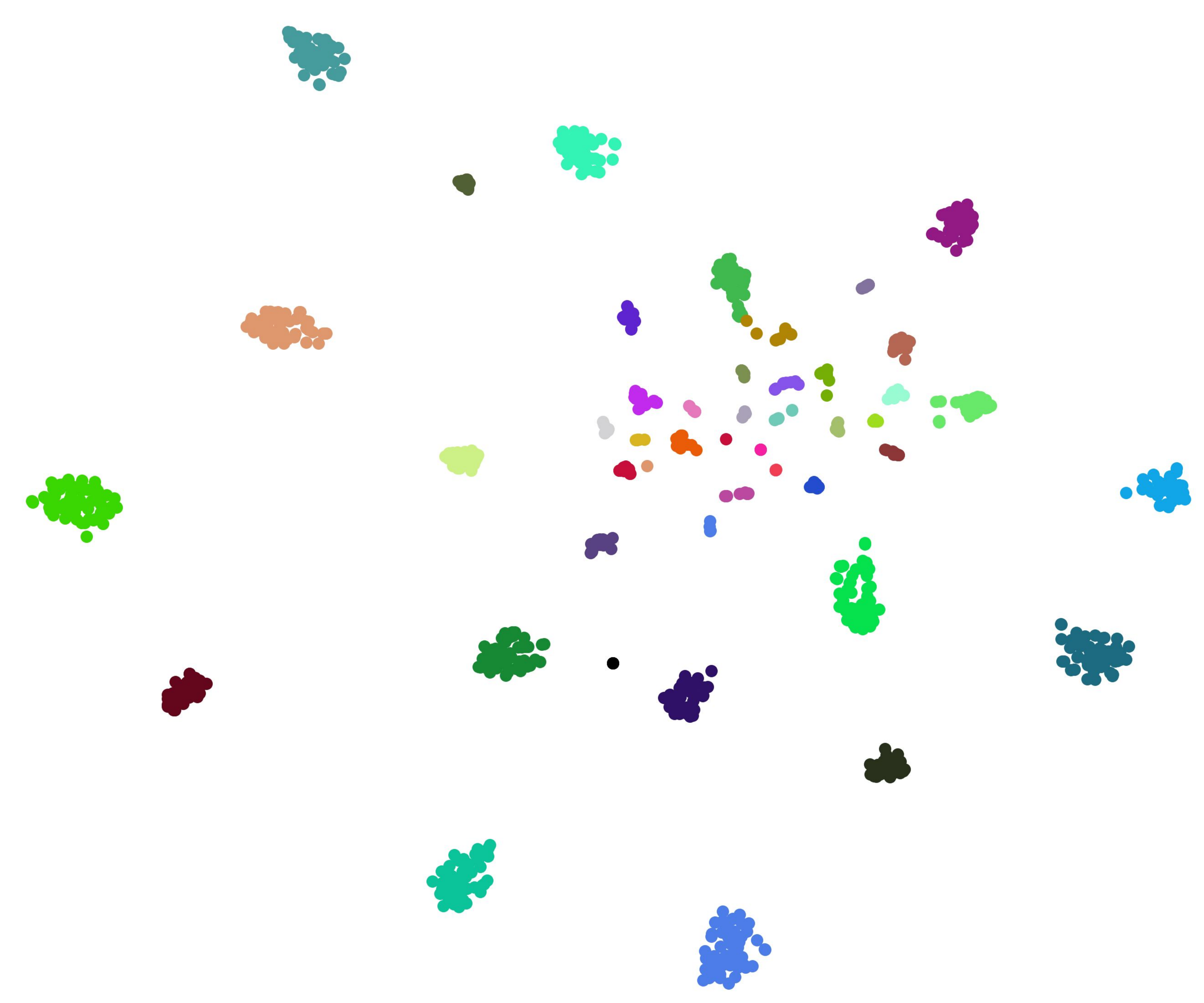}}
\caption{The t-SNE projection of the last hidden layer of target features (a) before adaptation and (b) after adaptation in the task of Syn.Sign$\rightarrow$GTSRB. Different colors represent different classes.}
\label{fig:tsne_projection_synthetic2real}
\end{figure}

To further demonstrate the effectiveness of our model, we visually inspect the generated images. Fig.~\ref{fig:class_conditional_generations_synthetic2real} shows the class-conditional generation on two tasks. In both scenarios, the generated images are consistent with the input labels and the style information is also encoded by the noise vector $z$. In addition, we visualize the distribution of target features before and after adaptation. As shown in Fig.~\ref{fig:tsne_projection_synthetic2real}, we use t-SNE~\cite{tSNE} to project the last hidden layer features onto the 2-D space in Syn.Sign$\rightarrow$GTSRB. The target instances are strongly clustered for each class after adaptation. These observations suggest that our model achieves accurate class-conditional generation in the target domain, which demonstrates superior model adaptation performance.

\begin{table}[t]
\small
\begin{center}
\resizebox{0.88\linewidth}{!}{
\begin{tabular}[\textwidth]{l c  c c c}
\hline
Method & \tabincell{c}{SVHN \\ $\downarrow$ \\ MNIST} &\tabincell{c}{MNIST \\ $\downarrow$ \\ USPS} & \tabincell{c}{USPS \\ $\downarrow$ \\ MNIST} & \tabincell{c}{MNIST \\ $\downarrow$ \\ MNIST-M}  \\
\hline
Source-Only                             & 68.1$\pm1.5$ & 85.3$\pm3.1$ & 71.0$\pm1.8$ & 50.3$\pm0.7$  \\
CMD~\cite{CMD}                          & 86.5         & -            & 86.3         & 85.5         \\
ADDA~\cite{ADDA}                        & 72.3         & 89.4         & 92.1         & 80.7         \\
CORAL~\cite{DeepCoral}                  & 89.5         & 81.7         & 96.5         & 81.6          \\
JDDA~\cite{JDDA}                        & 94.2         & -            & 96.7         & 88.4          \\
\hline
\multicolumn{5}{l}{\textit{Our Model Variants}} \\
\hline
w/o $\ell_{gen}$                    & - & - & - & - \\
w/ $\ell_{gen}$   & 97.9$\pm0.2$ & 94.5$\pm1.0$ & 98.2$\pm0.2$ & 91.8$\pm0.5$ \\
w/ $\ell_{gen}$, $\ell_{wReg}$   & 98.4$\pm0.2$ & 95.4$\pm0.3$ & 98.3$\pm0.1$ & 94.2$\pm0.3$ \\
Full Model                                           & \textbf{99.2$\pm0.1$} & \textbf{97.0$\pm0.2$} & \textbf{99.3$\pm0.1$} & \textbf{97.0$\pm0.1$} \\
\hline
\end{tabular}}
\end{center}
\caption{Ablation study on digit tasks with a small $C$ in JDDA~\cite{JDDA}. `-' denotes the results are not reported or do not converge.}
\label{table:ablation_study_on_digits}
\end{table}

\subsection{Ablation Study}

To demonstrate the robustness of the proposed method, we adopt a small classifier which is similar to LeNet used in JDDA~\cite{JDDA} for further evaluation. From Table~\ref{table:ablation_study_on_digits}, our full model still outperforms the Source-Only (baseline) by a large margin, which achieves about or more than 30\% improvement in most cases. Compared to the other unsupervised domain adaptation methods with the same classifier, our model achieves the best performance. For example, while JDDA reports an impressive performance (94.2\%) on the challenging SVHN$\rightarrow$MNIST task, our model surpasses this by about 5 percentage points. On the task MNIST$\rightarrow$MNIST-M, our model outperforms it by about 7 percentage points. These results demonstrate the effectiveness of our model.

\begin{figure}[t]
\centering 
\includegraphics[width=0.9\linewidth]{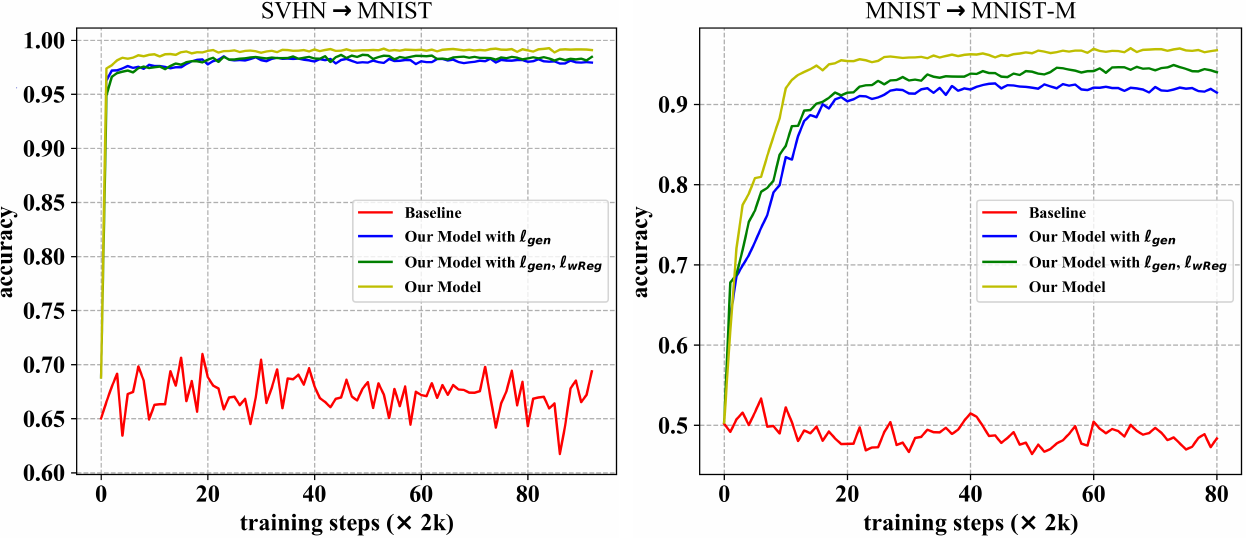}
\caption{Comparing the performance of our model variants on the task of (a) SVHN$\rightarrow$MNIST and (b) MNIST$\rightarrow$MNIST-M. The accuracy is computed on the target set w.r.t. the training steps.}
\label{fig:ablation_loss}
\end{figure}

To explore the capability of each component, we further compare the performance of our model variants by removing the corresponding modules or loss functions.

To evaluate the contribution of the generated images in improving the model adaptation, we first remove $\ell_{gen}$ in our 3C-GAN. From the last block of our model variants in Table~\ref{table:ablation_study_on_digits}, the model fails to converge without $\ell_{gen}$. We consider that the prediction model with only the proposed regularizations will hurt its discriminativity, due to the different distribution. Next we remove both regularizations $\ell_{wReg}$ and $\ell_{cluReg}$. The performance of our model with only $\ell_{gen}$ is significantly improved from the Source-Only model, as shown in the last third row of Table~\ref{table:ablation_study_on_digits}. These results imply that our 3C-GAN can achieve reliable class-conditional generation which facilitates the model adaptation performance. Detailed illustrations of accuracy curves during training on the task of SVHN$\rightarrow$MNIST and MNIST$\rightarrow$MNIST-M are presented in Fig.~\ref{fig:ablation_loss}. In both tasks, $\ell_{gen}$ is able to boost the accuracy of the baseline by a large margin, which is indicated by comparing the accuracy trends of the blue and the red curves in Fig.~\ref{fig:ablation_loss}.

To investigate the effectiveness of our proposed regularization terms, we disable $\ell_{cluReg}$ in Eq.~(\ref{eq:loss_C}) by setting $\lambda_{clu}=0$ during training. As shown in Table~\ref{table:ablation_study_on_digits}, the accuracy of our model by adding $\ell_{wReg}$ is further increased based on our model variant which only involves $\ell_{gen}$. We consider that the weight regularization not only prevents the model changing significantly but also inherits the knowledge in the pre-trained source model~\cite{TransferableFeatures}. Thus, it leads to more stable and better performance as indicated in Fig.~\ref{fig:ablation_loss} (Best viewed in color by comparing the blue and the green curves). Furthermore, by including the cluster regularization term $\ell_{cluReg}$, the performance of our full model can be consistently improved by around 1 to 3 percentage points on all the tasks. In particular, as shown in the last two rows of Table~\ref{table:ablation_study_on_digits}, the accuracy increases from 94.2\% to 97.0\% on MNIST$\rightarrow$MNIST-M, and 95.4\% to 97.0\% in MNIST$\rightarrow$USPS. It demonstrates that our clustering-based regularization can move the decision boundaries away from the dense data regions on the target domain, which increases the generalization of the prediction model.

Furthermore, we remove the smoothness constraint of Eq.~\ref{eq:l_smooth} to study the effect on adaptation performance. From Table~\ref{table:ablation_study_on_smoothness}, we observe that the accuracy dropped for the tasks A$\leftrightarrows$W and A$\leftrightarrows$D, which suggests that this constraint helps the conditional entropy estimation and improves the generalization performance.

\begin{table}[t]
\small
\begin{center}
\resizebox{0.90\linewidth}{!}{
\begin{tabular}[\textwidth]{l c c c c}
\hline
Method & A$\rightarrow$W & A$\rightarrow$D & D$\rightarrow$A & W$\rightarrow$A \\
\hline
w/o smoothness             & 93.4$\pm0.3$ & 91.0$\pm0.5$  & 74.0$\pm0.5$ & 77.3$\pm0.3$ \\
w/   smoothness             & 93.7$\pm0.2$ & 92.7$\pm0.4$  & 75.3$\pm0.5$  & 77.8$\pm0.1$ \\
\hline
\end{tabular}}
\end{center}
\caption{Ablation study to investigate effects of the smoothness.}
\label{table:ablation_study_on_smoothness}
\end{table}

\section{Conclusion}
In this paper, we propose a new model-based unsupervised domain adaptation method without source domain data. Since preparing a large amount of source data is inconvenient or even infeasible due to data privacy issues, our proposed method is more preferable for real-world applications. To this end, we propose 3C-GAN to bypass the dependence on source data. By incorporating generated images into the adaptation process, the prediction model and the generator can be mutually enhanced through collaborative learning. In addition, we introduce weight regularization and clustering-based regularization for stabilizing the training and further improving generalization performance on the target domain. We conduct extensive experiments on multiple domain adaptation benchmarks. Compared with recent data-based domain adaptation methods, our model achieves the best or comparable performance in the absence of source data, which demonstrates its effectiveness in a broad class of adaptation scenarios.

\noindent
\textbf{Acknowledgments.} This work was supported in part by the Research Grants Council of the Hong Kong Special Administration Region (Project No. CityU 11300715), in part by the National Natural Science Foundation of China (Project No. U1611461, 61722205, 61751205, 61572199), in part by City University of Hong Kong (Project No. 7005055), in part by the Natural Science Foundation of Guangdong Province (Project No. 2016A030310422, 2016A030308013), and in part by Fundamental Research Funds for the Central Universities (Project No. 2018ZD33).

{\small
\bibliographystyle{ieee_fullname}
\bibliography{egbib_abbre}
}

\end{document}